\documentclass[11pt, a4paper, logo, copyright]{main}

\usepackage{microtype}
\usepackage{graphicx}
\usepackage{subcaption}
\usepackage{booktabs}
\usepackage{hyperref}
\usepackage{multirow}
\usepackage{colortbl}
\usepackage{xcolor}
\usepackage[most]{tcolorbox}
\usepackage{fvextra}

\definecolor{firstcolor}{HTML}{F49767}  
\definecolor{secondcolor}{HTML}{F9BA86} 
\definecolor{thirdcolor}{HTML}{F5D5AC}  
\definecolor{goldcolor}{HTML}{F49767}
\definecolor{silvercolor}{HTML}{F5D5AC}

\usepackage{amsmath}
\usepackage{amssymb}
\usepackage{mathtools}
\usepackage{amsthm}

\theoremstyle{plain}

\theoremstyle{definition}

\theoremstyle{remark}

\usepackage[authoryear, sort&compress, round]{natbib}
\let\cite\citep
\usepackage{float}

\title{WebVR: Benchmarking Multimodal LLMs for WebPage Recreation from Videos via Human-Aligned Visual Rubrics}

\author[*]{
\textbf{Yuhong Dai}$^{1,*}$, \textbf{Yanlin Lai}$^{1,2,*}$, \textbf{Mitt Huang}$^{1,*,\dag,\heartsuit}$, \textbf{Hangyu Guo}$^{1,*}$, \\
\textbf{Dingming Li}$^{1}$, \textbf{Hongbo Peng}$^{1}$, \textbf{Haodong Li}$^{1}$, \textbf{Yingxiu Zhao}$^{1}$, \textbf{Haoran Lyu}$^{1}$, \\
\textbf{Zheng Ge}$^{1,\dag}$, \textbf{Xiangyu Zhang}$^{1}$, \textbf{Daxin Jiang}$^{1}$ \\
$^{1}$ StepFun \quad $^{2}$ Tsinghua University \\
\raisebox{-0.15em}{\includegraphics[height=0.9em]{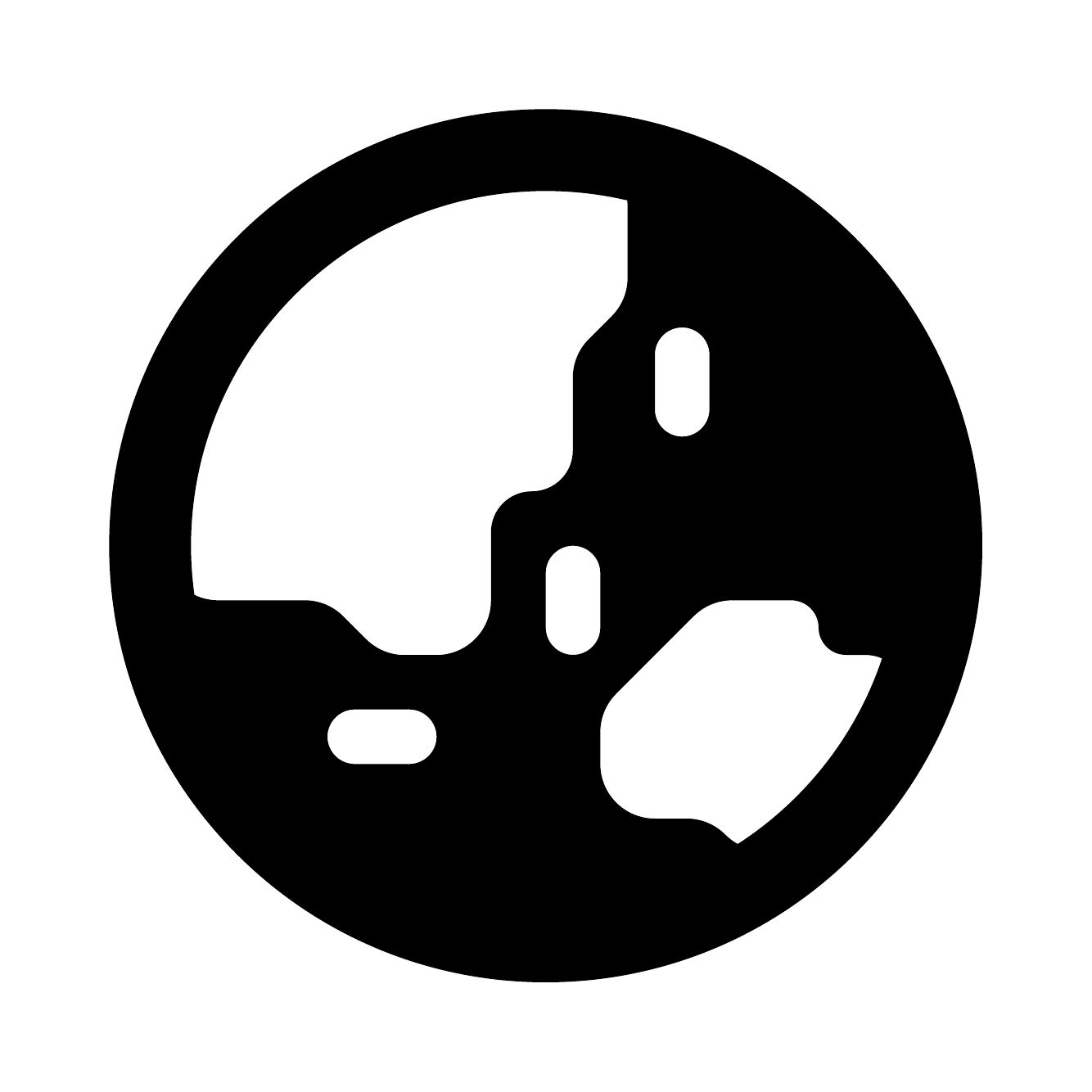}}\hspace{0.35em}\href{http://webvr-benchmark.github.io/}{Homepage}
\hspace{1.0em}
\raisebox{-0.15em}{\includegraphics[height=0.9em]{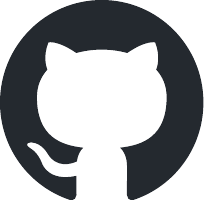}}\hspace{0.35em}\href{https://github.com/broalantaps/WebVR}{GitHub}
\hspace{1.0em}
\raisebox{-0.15em}{\includegraphics[height=0.9em]{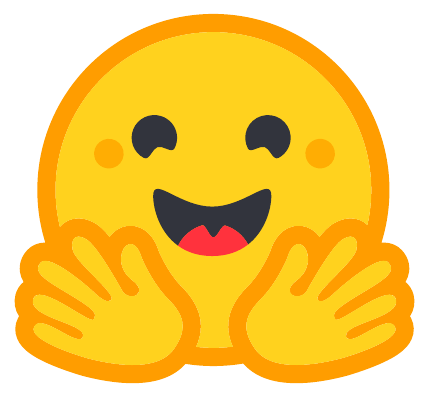}}\hspace{0.35em}\href{https://huggingface.co/datasets/BroAlanTaps/WebVR}{Hugging Face}
}

\setauthorfootnote{$*$ Equal Contribution \quad $\heartsuit$ Project Leader \quad $\dag$ Corresponding Author}

\reportnumber{} %

\renewcommand{\today}{}

\begin{abstract}

Existing web-generation benchmarks rely on text prompts or static screenshots as input. However, videos naturally convey richer signals such as interaction flow, transition timing, and motion continuity, which are essential for faithful webpage recreation. Despite this potential, video-conditioned webpage generation remains largely unexplored, with no dedicated benchmark for this task. To fill this gap, we introduce WebVR, a benchmark that evaluates whether MLLMs can faithfully recreate webpages from demonstration videos. WebVR contains 175 webpages across diverse categories, all constructed through a controlled synthesis pipeline rather than web crawling, ensuring varied and realistic demonstrations without overlap with existing online pages. We also design a fine-grained, human-aligned visual rubric that evaluates the generated webpages across multiple dimensions. Experiments on 19 models reveal substantial gaps in recreating fine-grained style and motion quality, while the rubric-based automatic evaluation achieves 96\% agreement with human preferences. We release the dataset, evaluation toolkit, and baseline results to support future research on video-to-webpage generation.

\end{abstract}

\begin{document}

\maketitle

\definecolor{colorfirst}{RGB}{252,141,89}
\definecolor{colorsecond}{RGB}{253,187,132}
\definecolor{colorthird}{RGB}{253,212,158}
\definecolor{colorfourth}{RGB}{254,232,200}
\definecolor{colorfifth}{RGB}{255,247,236}
\definecolor{myred}{RGB}{242,128,128}
\definecolor{mygreen}{RGB}{112,180,143}
\definecolor{myblue}{RGB}{210,225,255}
\definecolor{citypink}{RGB}{227,108,194}
\definecolor{cityblue}{RGB}{128,159,225}

\newcommand{\ph}[1]{\textcolor{black}{#1}}
\newcommand{\rankfirst}[0]{\cellcolor{colorfirst}}
\newcommand{\ranksecond}[0]{\cellcolor{colorsecond}}
\newcommand{\rankthird}[0]{\cellcolor{colorthird}}
\newcommand{\rankfourth}[0]{\cellcolor{colorfourth}}
\newcommand{\rankfifth}[0]{\cellcolor{colorfifth}}
\DeclareRobustCommand{\legendsquare}[1]{%
  \textcolor{#1}{\rule{2ex}{2ex}}%
}
\newcommand{\cmark}{\textcolor{mygreen}{\checkmark}}%
\newcommand{\xmark}{\textcolor{myred}{\times}}%

\section{Introduction}
\label{sec:intro}

MLLMs~\cite{lin2024video,bai2025qwen3,huang2026step3} are rapidly advancing toward end-to-end generation of executable artifacts from visual inputs. In web development, designers frequently provide screen recordings rather than fully specified design documents to communicate layout, interaction flow, and animation timing. These recordings naturally encode both static appearance and dynamic interactions, making them rich references for front-end implementation. Recent advances in MLLMs have made it feasible to convert such demonstration videos directly into executable code, substantially reducing development effort. However, the actual capabilities of current MLLMs in faithfully recreating webpages from video remain largely underexplored.

Existing benchmarks have advanced webpage evaluation but remain limited in two aspects. First, they primarily evaluate generation
from text prompts or static screenshots~\cite{xu2025web,lu2025webgen,sun2025fullfront,laurenccon2024unlocking,yun2024web2code}, leaving dynamic behaviors such as transitions and animations unevaluated. Even recent efforts that approximate dynamics via staged screenshots~\cite{zhang2025artifactsbench} provide only sparse temporal evidence, failing to capture the fluid motion essential to modern web
interfaces. Second, existing evaluation protocols rely on coarse-grained or structure-oriented criteria, lacking a principled rubric for assessing fine-grained visual and interaction fidelity.

\begin{figure}[t]
  \centering
  \includegraphics[width=\linewidth]{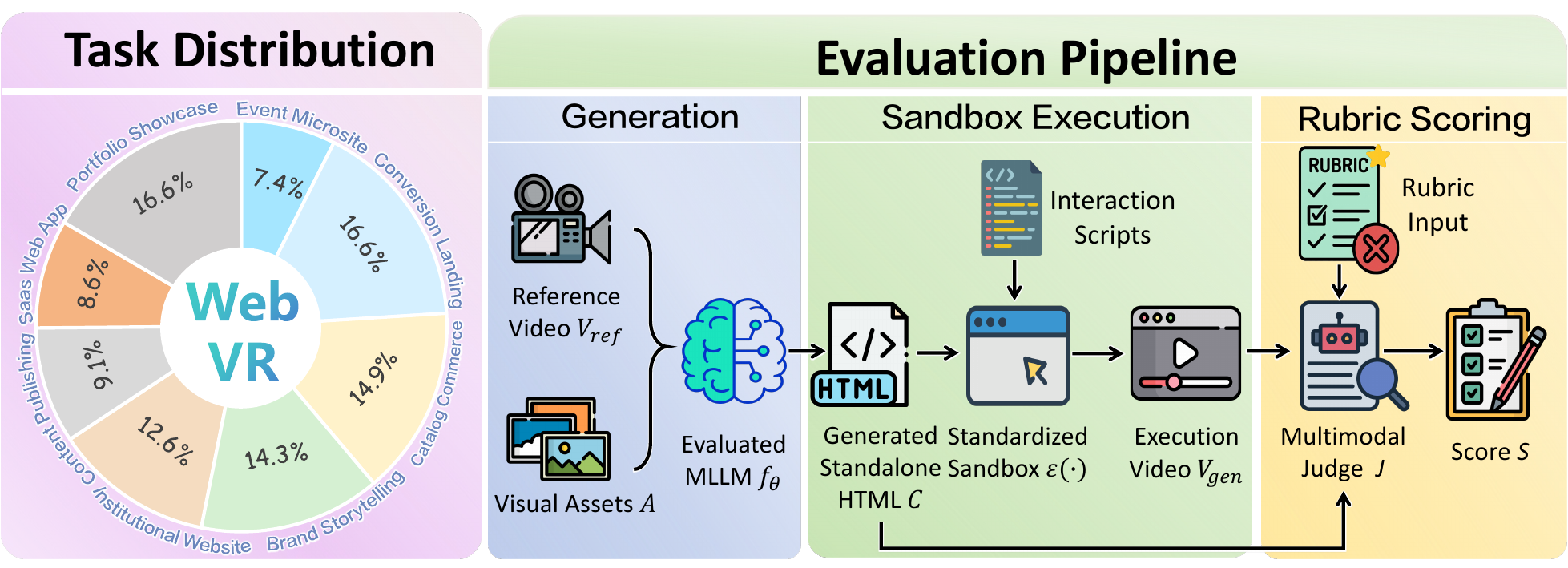}
  \caption{Overview of the WebVR benchmark. The left panel illustrates the task distribution across various webpage categories, while the right panel details the automated evaluation pipeline, which executes generated code in a standardized sandbox and scores it against human-aligned visual rubrics.}
  \label{fig:evalpipeline}
\end{figure}

To fill these gaps, we introduce \textbf{WebVR}, the first benchmark for video-to-webpage generation. WebVR evaluates whether MLLMs can transform demonstration videos into fully functional front-end implementations with high visual and interactive fidelity. It contains 175 webpages across diverse categories such as e-commerce, portfolio, landing page, entertainment, and education. All webpages are constructed through a controlled synthesis pipeline rather than web crawling, ensuring varied and realistic demonstrations without overlap with existing online pages. To support fair evaluation, WebVR provides visual assets alongside each task, so that models focus on faithfully reconstructing layout and interactions rather than sourcing suitable images. We also design a fine-grained, human-aligned visual rubric that evaluates the generated webpages across multiple dimensions including layout structure, color and typography, component completeness, animation quality, and interaction correctness. The generated code is rendered and recorded in a standardized sandbox, and an MLLM-based judge scores the execution video against the reference video under this rubric, enabling scalable, reproducible, and interpretable assessment. An overview of WebVR is shown in Fig.~\ref{fig:evalpipeline}. Experiments on 19 models reveal substantial gaps in recreating fine-grained style and motion quality, while the rubric-based automatic evaluation achieves 96\% agreement with human preferences.

The contributions of this work are as follows:

\begin{itemize}
    \item \textbf{Video-to-Webpage Benchmark.} We introduce \textbf{WebVR}, the first benchmark for video-to-webpage generation. It covers 175 visually rich webpages across diverse categories, constructed through a controlled synthesis pipeline with paired visual assets for fair evaluation.
    \item \textbf{Video-Based Evaluation Protocol.} We propose an evaluation protocol that renders the generated front-end code in a standardized sandbox and records execution videos, enabling reproducible assessment of both static appearance and dynamic interactions beyond static snapshots.
    \item \textbf{Human-Aligned Visual Rubric.} We design a fine-grained, human-aligned visual rubric that guides an MLLM-based judge~\cite{chen2024mllm} to score generated webpages across multiple dimensions, producing interpretable, dimension-level feedback and reliable model rankings that achieve 96\% agreement with human preferences.
\end{itemize}

\section{Related Work}
\label{sec:related_work}
\begin{table*}[t]
\centering
\caption{Feature Comparison of Existing Webpage Benchmarks (AP: Asset-Provided; GL: Granularity-Level)}
\label{tab:comparison}
\resizebox{\textwidth}{!}{
    \begin{tabular}{lccccc} 
    \toprule
    \textbf{Benchmark} & \textbf{Input Modality} & \textbf{Data Source} & \textbf{Evaluation Modality} & \textbf{AP} & \textbf{GL} \\
    \midrule
    Web2Code       & Image & Synthetic     & Image         & $\times$ & Low  \\
    WebBench       & Text  & Human-Written & Code          & $\times$ & Mid  \\
    WebUIBench     & Image & Real-World    & Code \& Image & $\times$ & Mid  \\
    FullFront      & Image & Synthetic     & Code \& Image & $\times$ & Mid  \\
    ArtifactsBench & Text  & Synthetic     & Code \& Image & $\times$ & High \\
    \midrule
    \rowcolor[HTML]{EFEFEF}
    \textbf{WebVR (Ours)} & \textbf{Video} & \textbf{Synthetic} & \textbf{Code \& Video} & \checkmark \textbf{(Fair)} & \textbf{High} \\
    \bottomrule
    \end{tabular}
}
\end{table*}


\noindent\textbf{Webpage Generation.}
Recent progress in LLMs~\cite{naveed2025comprehensive, dai2025pretraining} has enabled translating natural language specifications into executable web code. Several benchmarks evaluate this capability from different input modalities. Text-based benchmarks such as ArtifactsBench~\cite{zhang2025artifactsbench} and Interaction2Code~\cite{xiao2025interaction2code} approximate dynamic behaviors through staged screenshots, but this sparse sampling fails to capture fine-grained motion patterns, animation timing, and continuous transitions. WebRRSBench~\cite{liu2025benchmarking} broadens evaluation to reasoning, robustness, and safety, yet does not address dynamic visual fidelity. Image-based benchmarks including WebSight~\cite{laurenccon2024unlocking}, Web2Code~\cite{yun2024web2code}, DesignBench~\cite{xiao2026designbenchcomprehensivebenchmarkmllmbased}, and WebUIBench~\cite{lin2025webuibench} advance visual fidelity by translating screenshots or UI slices into code. However, they remain grounded in static inputs and provide no supervision over temporal dynamics such as animation pacing and user-triggered state transitions. In contrast, video demonstrations inherently encode rich spatio-temporal signals including motion trajectories, scrolling behaviors, and hover feedback. To the best of our knowledge, no existing benchmark systematically evaluates the recovery of these fine-grained dynamic behaviors from video demonstrations, which is the focus of this work.
We summarize the key differences between existing webpage benchmarks
and our WebVR in Table~\ref{tab:comparison}.

\noindent\textbf{MLLM as Judge.}
Traditional DOM or pixel-based metrics~\cite{radford2021learning} often overlook high-level semantics and interaction flows. Recent work has explored using MLLMs as evaluators: WebDevJudge~\cite{li2026webdevjudgeevaluatingmllmscritiques} and MLLM as a UI Judge~\cite{luera2025mllmuijudgebenchmarking,li2026gebench} assess web development quality and human perception of UIs, respectively, while WebCoderBench~\cite{liu2026webcoderbench} introduces 24 fine-grained metrics combining rule-based and LLM-as-a-judge paradigms. The emergence of LLM-as-a-Judge~\cite{10.5555/3666122.3668142}, together with advances in multimodal models such as Qwen3-VL~\cite{bai2025qwen3} and Kimi-K2.5~\cite{team2026kimi}, has made it feasible to assess visual fidelity and layout coherence directly from rendered webpages. However, prior studies report challenges including hallucination~\cite{wang-etal-2024-large-language-models-fair}, prompt sensitivity, and limited awareness of engineering quality. WebVR addresses these by employing a fine-grained visual rubric aligned with human preferences, guiding the judge to produce reliable and interpretable scores.

\section{Method}
\label{sec:method}

\begin{figure}[t]
  \centering
  \includegraphics[width=0.85\linewidth]{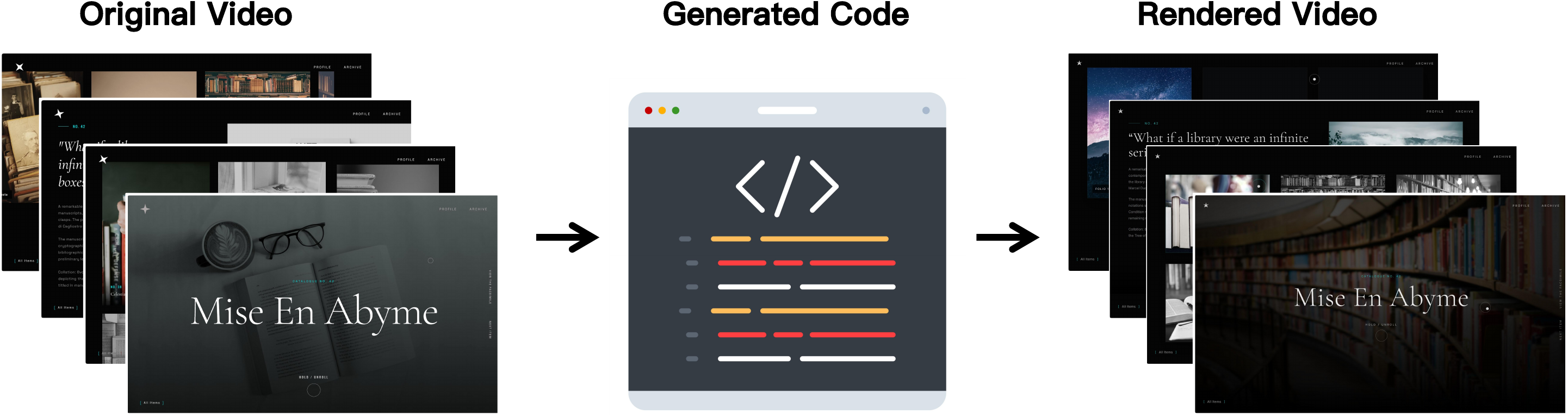}
  \caption{A case study illustrating the conversion from an original video into generated code and its rendered video output.}
  \label{fig:case}
\end{figure}

\subsection{Task Definition}
\label{sec:task_def}
We define the task of video-conditioned webpage recreation. Given a reference screen-recording video $V_{ref}$ that demonstrates the target webpage's layout, visual style, and interactive behaviors, along with a set of visual assets $A = \{a_1, a_2, \dots, a_m\}$ (e.g., images and icons), an MLLM $f_\theta$ is required to generate a standalone, executable HTML document:
\begin{equation}
C = f_\theta(V_{ref}, A)
\end{equation}
Unlike screenshot-to-code tasks that focus on static visual cloning, this task requires the model to accurately reproduce both spatial layout and dynamic interactive behaviors demonstrated in the video.

\noindent\textbf{Dynamic Execution and Rendering.} To evaluate the true user-facing output, we introduce a standardized sandbox environment $\mathcal{E}(\cdot)$ that executes the generated HTML $C$ under a fixed interaction script and records the resulting screen-capture video:
\begin{equation}
V_{gen} = \mathcal{E}(C)
\end{equation}
This enables direct visual comparison between $V_{gen}$ and $V_{ref}$, capturing both static appearance and dynamic state transitions. Fig.\ref{fig:case} provides a visual case study of this end-to-end process, illustrating the conversion from the original reference video to the generated code and its rendered output.

\begin{figure}[h]
  \centering
  \includegraphics[width=\linewidth]{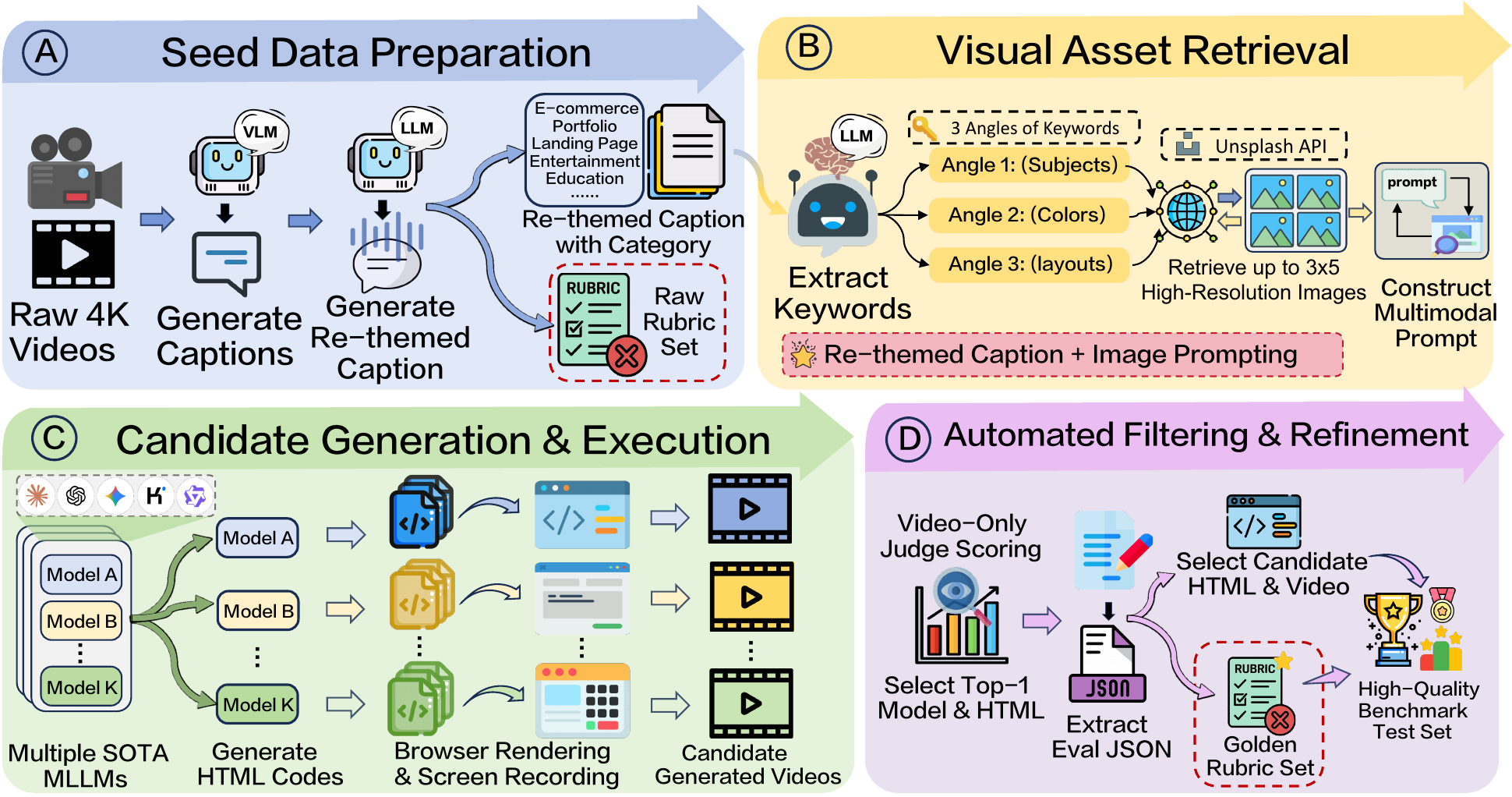}
  \caption{The WebVR data synthesis pipeline. The process consists of four stages: (A) Seed Data Preparation via semantic re-theming, (B) Visual Asset Retrieval to ground specifications, (C) Candidate Generation and Execution using multiple MLLMs, and (D) Automated Filtering and Refinement to construct the final high-quality benchmark set.}
  \label{fig:datapipeline}
\end{figure}

\noindent\textbf{Evaluation via Visual Rubrics.} A perfect recreation would yield $V_{gen} \approx V_{ref}$. However, direct pixel-wise or frame-by-frame comparison is brittle for generative tasks, as it penalizes minor yet acceptable variations in animation easing or rendering differences.
Instead, we decompose visual and interactive consistency into a set of fine-grained, atomic rubric criteria $R = \{r_1, r_2, \dots, r_K\}$, uniquely defined for each reference video $V_{ref}$. Each criterion $r_k$ is a binary verification condition targeting a single visual property, such as whether a specific element exists, whether its alignment is correct, or whether a hover animation triggers as expected.
An MLLM-based judge $J$ evaluates the generated output against the full rubric $R$ in a single inference pass. The judge takes both the execution video $V_{gen}$ and the generated HTML $C$ as input, and predicts the binary satisfaction status for each criterion, producing $\hat{Y}=\{\hat{y}_1,\hat{y}_2,\dots,\hat{y}_K\}$ where $\hat{y}_k\in\{0,1\}$. The overall recreation quality score is computed as:
\begin{equation}
S=\frac{1}{K}\sum_{k=1}^{K}\hat{y}_k
\end{equation}
This formulation shifts webpage evaluation from coarse structural or textual metrics to a visually grounded, temporally aligned verification process.

\subsection{Seed Data Preparation}
\label{sec:webvr_curation}

A primary challenge in benchmarking modern MLLMs is data contamination, as these models have likely been exposed to a vast majority of existing web content during pre-training. As illustrated in Fig. \ref{fig:datapipeline}, to ensure that evaluation instances are unseen by the models, we construct each WebVR instance through a multi-stage synthesis process: (i) collecting real-world webpage demonstration videos, (ii) translating them into structured captions, and (iii) re-theming each caption into a fictional but structurally equivalent specification. The collected videos are used only to derive captions and specifications; benchmark input videos are produced later via candidate generation and execution.

\noindent\textbf{Video Collection.} We collect screen-recorded website showcase videos from design-gallery platforms, including Landing Love, Godly, and Lapa. We retain high-resolution recordings when available to preserve fine-grained typography and motion cues.

\noindent\textbf{Structured Captioning.} For each collected video, we use an MLLM to produce a structured caption $T$ that abstracts the page into an explicit design specification covering global aesthetics, section layouts, reusable components, and interaction logic. Unlike brittle pixel-level transcription, this intermediate representation enables semantic re-theming and serves as a stable input for rubric generation and reference synthesis in later stages.

\noindent\textbf{Semantic Re-theming.} Since MLLMs may have encountered real webpages during their training, directly using them as benchmark instances risks data leakage. To prevent this, we rewrite $T$ into a fictional specification $\tilde{T}$ that preserves the original layout and interaction logic but replaces all semantic content, including domain, entities, copy, and brand identifiers. Motion descriptions are kept consistent to retain temporal behaviors.


\noindent\textbf{Caption categorization.}
For diversity analysis and potential balancing, we optionally classify each re-themed caption into one of eight coarse categories in our current pipeline, including \textbf{Brand Storytelling}, which focuses on conveying a brand’s narrative and identity; \textbf{Catalog Commerce}, emphasizing product listings and transactional features; \textbf{Conversion Landing}, designed to drive specific user actions such as sign-ups or purchases; \textbf{Content Publishing}, centered on articles, blogs, or media content; \textbf{Event Microsite}, highlighting temporary or campaign-specific events; \textbf{Institutional Website}, representing organizational, corporate, or educational sites; \textbf{SaaS Web App}, covering interactive software-as-a-service platforms; and \textbf{Portfolio Showcase}, which presents creative work or professional portfolios in a visually compelling format.

\subsection{Visual Asset Retrieval and Candidate Generation}
\label{sec:webvr_assets}

\noindent\textbf{Visual Asset Retrieval.} Modern webpages rely heavily on high-quality imagery. To reduce ambiguity during code generation, we ground each re-themed caption with a small set of public images retrieved from Unsplash. For each sample, an LLM generates three sets of keywords from different perspectives (e.g., subjects, colors, layouts) and queries the Unsplash API, returning up to five images per set. The retrieved images and their URLs are provided to the model in a fixed order, enabling deterministic asset referencing and placement.

\noindent\textbf{Candidate Generation and Execution.} Given a re-themed caption $\tilde{T}$, its grounded assets $A$, and the generated rubric $R$ (described in Sec.~\ref{sec:webvr_rubric}), we produce a high-quality reference execution video $V_{ref}$ to serve as the benchmark input. For each sample, we prompt five code-generation models to each produce a standalone HTML implementation $C_j$ conditioned on $(\tilde{T}, A)$, yielding five candidates $\{C_j\}_{j=1}^{5}$. Each candidate is rendered using the standardized executor $\mathcal{E}(\cdot)$ defined in Sec.~\ref{sec:task_def}, producing a candidate execution video $V_j = \mathcal{E}(C_j)$.




\begin{figure}[t]
  \centering
  \captionsetup[subfigure]{justification=centering}
  \begin{subfigure}{0.48\linewidth}
    \centering
    \includegraphics[width=\linewidth]{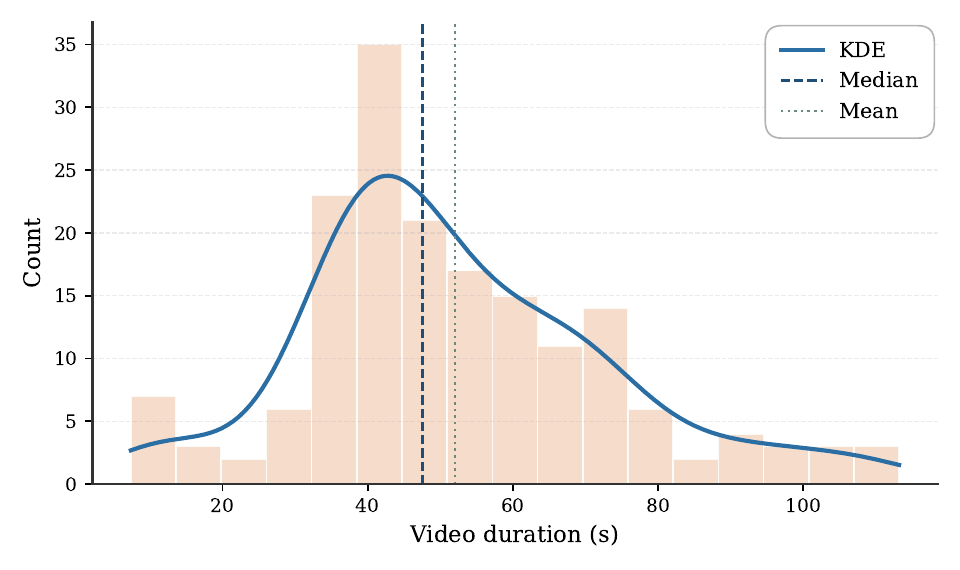}
    \caption{Video duration distribution.}
    \label{fig:video_dur}
  \end{subfigure}
  \hfill
  \begin{subfigure}{0.48\linewidth}
    \centering
    \includegraphics[width=\linewidth]{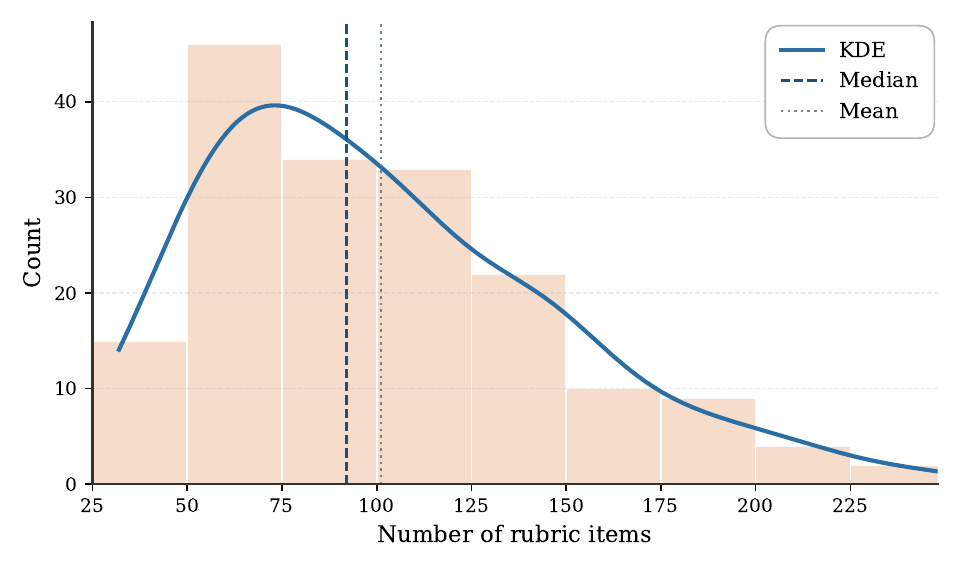}
    \caption{Rubric item count distribution.}
    \label{fig:rubric_cnt}
  \end{subfigure}

  \caption{Statistics of the WebVR benchmark dataset, showing the distributions of (a) reference video durations (in seconds) and (b) the number of visual rubric items per instance.}
  \label{fig:overall_label}
\end{figure}

\subsection{Automated Filtering and Refinement}
\label{sec:webvr_rubric}
In this section, we detail the automated synthesis and application of visual evaluation rubrics. Specifically, we describe how fine-grained rubrics are generated from the re-themed captions and subsequently utilized to filter candidate implementations, score visual fidelity, and refine the final benchmark references.

\noindent\textbf{Taxonomy of Web Design Dimensions.} We organize rubric criteria into four orthogonal dimensions covering user perception from global aesthetics to local interactions:
\begin{itemize}
  \item \textbf{Global Aesthetics (GA):} overall color mood, typographic character, and coherence of the visual language.
  \item \textbf{Navigation and Footer (NF):} structure and visible states of persistent UI components such as header menus and footers.
  \item \textbf{Section-Specific Layouts (SSL):} section-level layout, grid structure, hierarchy, alignment, and spacing for each distinct section.
  \item \textbf{Interaction and Motion (IM):} dynamic behaviors such as hover feedback, scroll-triggered animations, and state transitions.
\end{itemize}
This taxonomy supports dimension-level diagnostics and serves as the organizing scaffold for rubric synthesis.

\noindent\textbf{Automated Rubric Synthesis.} We use a pool of rubric-generator models to produce a visual verification rubric $R=\{r_k\}_{k=1}^{K}$ from the re-themed caption $\tilde{T}$. Each $r_k$ is a binary check phrased in terms of rendered visual evidence and tagged with one of the dimensions above. To ensure robustness and attributability, we enforce three generation rules:
\begin{itemize}
  \item \textbf{Visual proof only:} criteria describe visible properties of the rendered page (e.g., presence, alignment, color tone), not HTML tags, CSS class names, or implementation details.
  \item \textbf{Extreme atomicity:} each criterion checks exactly one attribute and avoids conjunctions, preventing ambiguous partial satisfaction.
  \item \textbf{Decomposition strategy:} for each key element in $\tilde{T}$, criteria are decomposed along existence/content, layout/position, and style/appearance.
\end{itemize}

\noindent\textbf{Rubric-Based Selection.} We score each candidate using the judge-based rubric evaluation defined in Sec.~\ref{sec:task_def}. For this selection step, the judge takes only the rendered video $V_j$ and the rubric $R$ as input (without HTML). We select the highest-scoring candidate:
\begin{equation}
  j^{*} = \arg\max_{j \in \{1,\dots,5\}} S(V_j, R), \quad V_{ref} = V_{j^{*}}, \quad C_{ref} = C_{j^{*}}.
\end{equation}
To construct a model-balanced reference set, we group samples by the winning generator model and take the top 50 samples per model by $S(V_{ref}, R)$ (250 total), then remove samples with $S(V_{ref}, R) < 0.5$, yielding 175 benchmark instances.

\noindent\textbf{Rubric Pruning.} Even the best candidate may not satisfy every rubric item due to model limitations or rendering differences. Since the benchmark input is the reference video $V_{ref}$ itself, the evaluation rubric must be satisfiable given that input. We therefore remove any criterion with $\hat{y}_k=0$ from the selection stage, producing a pruned rubric $R^{*} \subseteq R$. Additionally, during benchmark construction, we filter out insignificant hover elements from the reference to focus on meaningful interactions; during evaluation, all hover elements in the model-generated HTML are tested without filtering.

Each benchmark instance consists of the reference video $V_{ref}$, the grounded asset set $A$, and the pruned rubric $R^{*}$. The pruned rubric sizes range from 32 to 248 criteria per sample (Fig.~\ref{fig:rubric_cnt}), and reference video durations are summarized in Fig.~\ref{fig:video_dur}.

\section{Experiments}
\label{sec:experiment}

\subsection{Experimental Setup}
In our experimental setting, GPT and Claude series models use uniformly sampled 32 frames as video input. All other models follow their official API pipelines for video processing, with a unified sampling rate of 2 FPS. For inference hyperparameters, we adopt the recommended default or best-performing configurations provided in the official documentation of each model to ensure fairness and reproducibility. 

The evaluation sandbox environment utilizes the Chromium engine with a resolution of 2560x1440 at 30 FPS to test page scrolling and hover effects. To simulate authentic user needs, we filtered out potentially insignificant hover elements during benchmark construction. During evaluation, we test all hover elements within the model-generated HTML.

\definecolor{bestcolor}{RGB}{252,243,229}
\begin{table*}[t]
\centering
\large
\setlength{\tabcolsep}{6.5pt}
\renewcommand{\arraystretch}{1.2}

\caption{Results of the WebVR Benchmark. Joint evaluation based on code and rendered video, using Kimi-K2.5 as the evaluator. The best results are marked in \colorbox{bestcolor}{\textbf{orange}}, and the second best results are marked with an \underline{underline}.}
\begin{tabular}{c c c c c c}
\toprule
\textbf{Model} 
& \textbf{GA} 
& \textbf{NF} 
& \textbf{SSL} 
& \textbf{IM} 
& \textbf{Overall} \\
\midrule

\rowcolor{blue!5}
\multicolumn{6}{c}{\textbf{Open-source Models}} \\
GLM-4.6V 
& 22.78 
& 15.42 
& 7.17 
& 14.35 
& 11.42 \\
Qwen3-VL-30B-A3B-Instruct 
& 33.33 
& 34.87
& 17.71
& 12.67
& 21.44 \\
Qwen3-VL-30B-A3B-Thinking 
& 53.38
& 60.47 
& 33.49 
& 20.22 
& 37.69  \\
Qwen3-VL-235B-A22B-Instruct 
& 51.06
& 52.65 
& 40.09 
& 22.12
& 40.71 \\
Qwen3-VL-235B-A22B-Thinking 
& 61.20 
& 68.04 
& 43.11 
& 29.30 
& 46.80 \\
Qwen3.5-397B-A17B 
	& 80.46 
	& 76.62 
	& 58.81 & 41.96 
	& 61.33 \\
Kimi-K2.5
& 87.44 & \underline{89.21} & \cellcolor{bestcolor}\textbf{79.26} & \cellcolor{bestcolor}\textbf{60.10} & \cellcolor{bestcolor}\textbf{79.14} \\

\midrule

\rowcolor{blue!5}
\multicolumn{6}{c}{\textbf{Closed-source Models}} \\
GPT-4.1
& 61.91 & 64.42 & 42.70 & 26.71 & 45.85 \\
GPT-5.2-Thinking
& \cellcolor{bestcolor}\textbf{89.76} & 89.08 & 77.27 & \underline{59.97} & 77.93 \\

Gemini-2.5-Flash 
& 71.95 & 70.90 & 51.52 & 39.76 & 55.62 \\
Gemini-2.5-Pro 
& 78.59 & 80.17 & 59.56 & 48.66 & 63.09 \\
Gemini-3.0-Flash 
& 84.05 & 85.19 & 67.74 & 48.43 & 69.49 \\
Gemini-3.0-Pro 
& 80.84 & 81.79 & 66.31 & 46.86 & 67.32 \\
Gemini-3.1-Pro-Preview
& \underline{88.30} & 87.29 & 77.09 & 56.50 & 76.69 \\

Claude-Sonnet-3.7
& 76.38 & 80.54 & 59.26 & 37.38 & 61.21 \\
Claude-Sonnet-4.6
& 87.16 & \cellcolor{bestcolor}\textbf{89.37} & \underline{78.87} & 59.06 & \underline{78.49} \\
Claude-Opus-4.6
& 87.66 & 87.98 & 78.60 & 54.33 & 77.33 \\

Seed-1.8
& 75.06 & 77.95 & 62.21 & 36.33 & 61.98 \\
Seed-2.0-Pro
& 82.88 & 86.27 & 73.35 & 45.88 & 71.88 \\

\bottomrule
\end{tabular}

\label{tab:main_results}
\end{table*}

\begin{figure}[t]
  \centering
  \begin{subfigure}{0.32\linewidth}
    \centering
    \includegraphics[width=\linewidth]{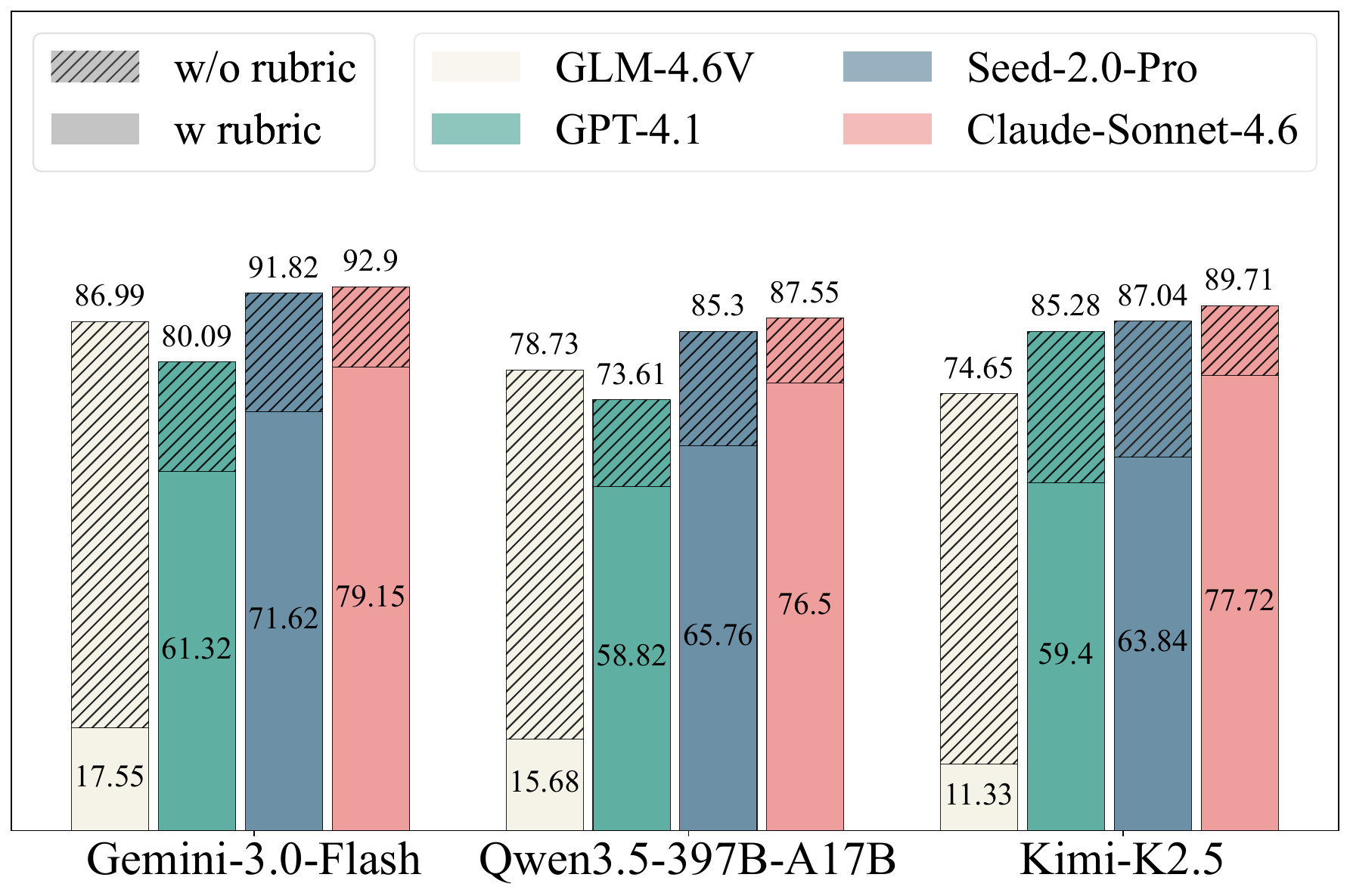}
    \caption{Code Only}
  \end{subfigure}
  \hfill
  \begin{subfigure}{0.32\linewidth}
    \centering
    \includegraphics[width=\linewidth]{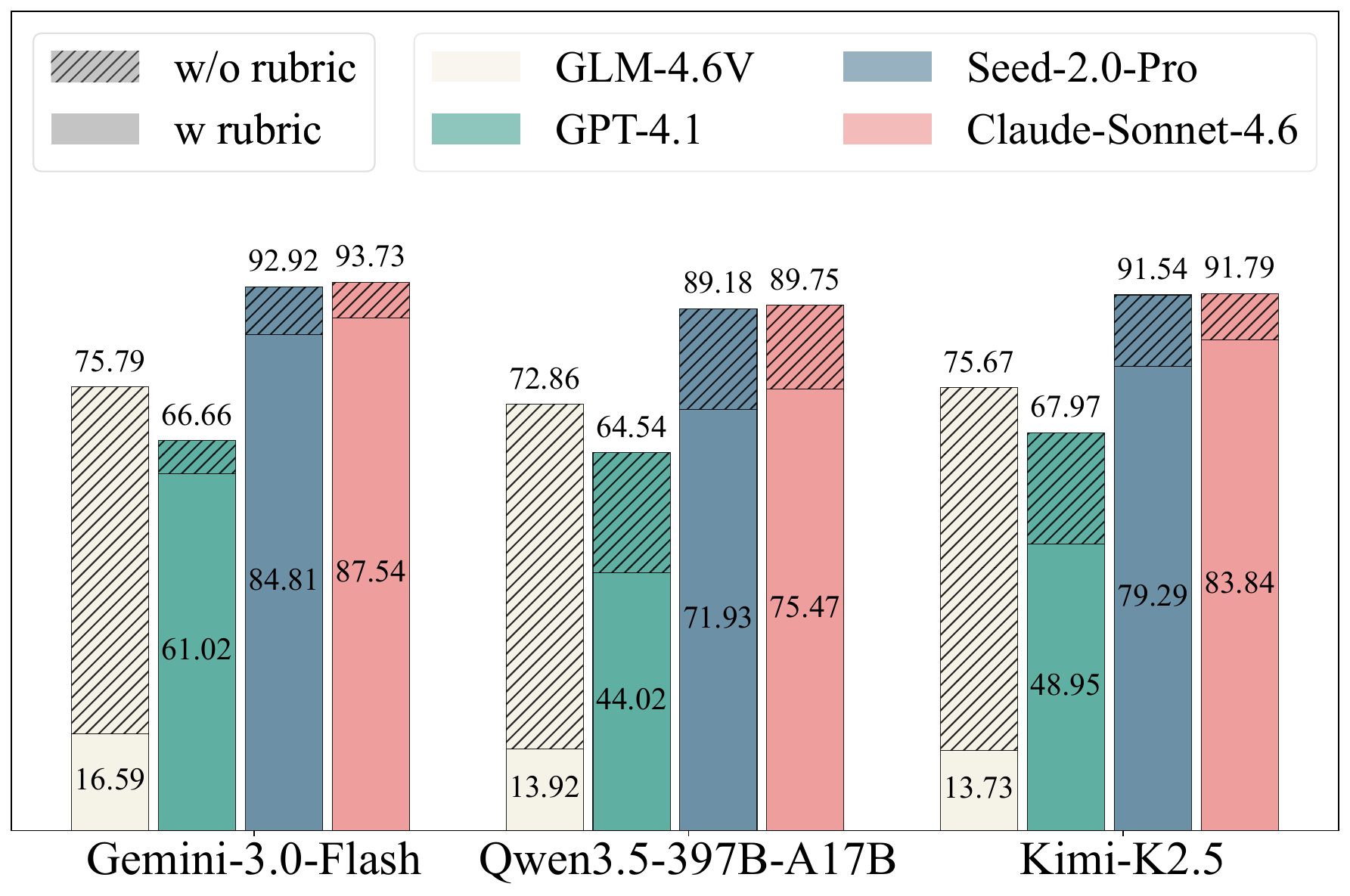}
    \caption{Video Only}
  \end{subfigure}
  \hfill
  \begin{subfigure}{0.32\linewidth}
    \centering
    \includegraphics[width=\linewidth]{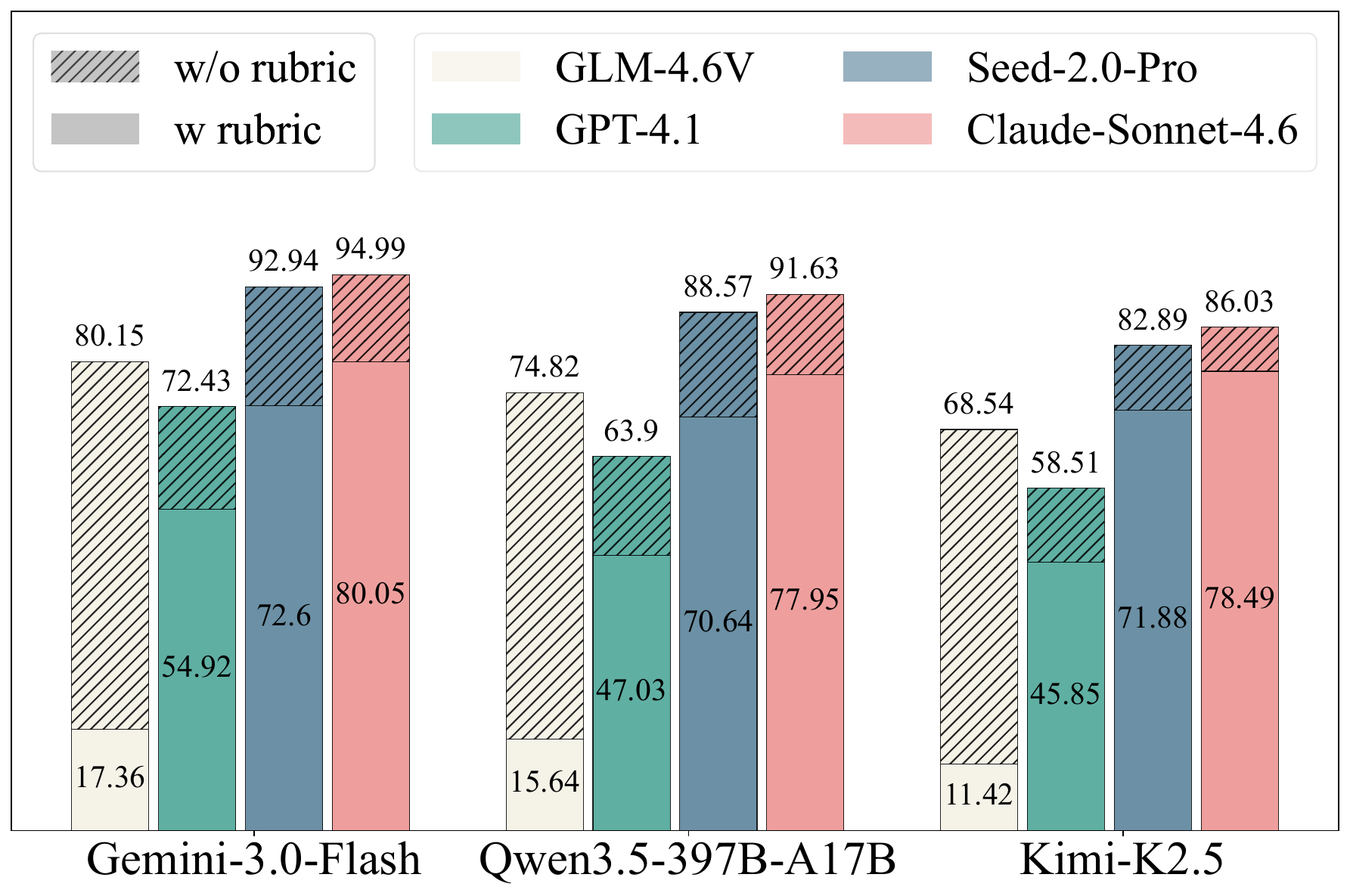}
    \caption{Code and Video}
  \end{subfigure}
  
  
  \caption{Hashed bars represent scores assigned without the visual rubric, while solid bars represent scores with the rubric applied. Note the inflated and compressed scoring distribution when the rubric is absent.}
  \label{fig:rubric_impact}
\end{figure}
\subsection{Main results}
Table~\ref{tab:main_results} reports the main results on WebVR, including dimension-level rubric fulfillment scores for Global Aesthetics (GA), Navigation and Footer (NF), Section-Specific Layouts (SSL), and Interaction and Motion (IM), as well as the overall score.

Overall, Kimi-K2.5 achieves the best overall score (79.14), closely followed by Claude-Sonnet-4.6 (78.49) and GPT-5.2-Thinking (77.93).
Notably, the best-performing open-source model is competitive with, and slightly surpasses, the best closed-source alternatives on this benchmark.

Across dimensions, the top models reach high scores on GA/NF (e.g., 89.76 GA for GPT-5.2-Thinking and 89.37 NF for Claude-Sonnet-4.6), suggesting that capturing global style cues and persistent UI components is increasingly feasible.
In contrast, Interaction and Motion (IM) remains the bottleneck: even the strongest models peak around 60 (60.10 for Kimi-K2.5 and 59.97 for GPT-5.2-Thinking).
Section-specific layout reconstruction is also challenging compared to GA/NF, with the highest SSL score at 79.26 (Kimi-K2.5) but substantially lower SSL scores for many mid-tier models.

We also observe clear improvements from scaling and newer generations.
Within the Qwen family, both model size scaling and test-time scaling matter. The reasoning-enhanced Thinking edition consistently outperforms the standard Instruct edition: Qwen3-VL-30B improves from 21.44 (Qwen3-VL-30B-A3B-Instruct) to 37.69 (Qwen3-VL-30B-A3B-Thinking), and Qwen3-VL-235B improves from 40.71 (Qwen3-VL-235B-A22B-Instruct) to 46.80 (Qwen3-VL-235B-A22B-Thinking). Scaling to Qwen3.5-397B-A17B further pushes performance to 61.33.
Similarly, Gemini improves from 55.62 (Gemini-2.5-Flash) to 76.69 (Gemini-3.1-Pro-Preview), and Seed improves from 61.98 (Seed-1.8) to 71.88 (Seed-2.0-Pro).

However, these gains are unevenly distributed across evaluation dimensions. Across all 19 models, we observe a consistent difficulty ordering: GA $>$ NF $>$ SSL $>$ IM, with average scores of 72.57, 72.09, 43.27, and 38.44, respectively. This reveals a steep drop from global aesthetics to fine-grained interactions. Even for top-tier models, the IM--GA gap exceeds 27 points (e.g., 27.34 for Kimi-K2.5, 29.79 for GPT-5.2-Thinking), indicating that current MLLMs can extract high-level visual style from video frames but struggle to translate temporal cues into executable interaction logic. This difficulty gradient is further corroborated by the Instruct/Thinking comparison: explicit deliberation primarily boosts the easier dimensions. For Qwen3-VL-30B, thinking yields a +16.25 overall gain, driven by large improvements on GA/NF (+20.05/+25.60), while for Qwen3-VL-235B the pattern repeats at a smaller scale (+6.09 overall, again led by GA/NF). Nevertheless, IM remains stubbornly low even with thinking enabled (20.22 and 29.30 for the two thinking variants), confirming that better ``planning'' alone does not resolve the temporal interaction bottleneck.

\begin{table*}[t]
\centering
\setlength{\tabcolsep}{7pt}
\renewcommand{\arraystretch}{1.08}
\caption{Agreement rates (\%) between automated MLLM judges and human experts. Evaluated on 50 sampled instances, the table contrasts judge reliability with and without the visual rubric across different input modalities.}
\label{tab:human_pref}
\begin{tabular}{c|ccc|c}
\toprule
\textbf{\small Evaluation Type} \quad & \textbf{\small Gemini-3.0-Flash} \quad & \textbf{\small Qwen3.5-397B-A17B} \quad & \textbf{\small Kimi-K2.5} \quad & \textbf{\small Average}\\
\midrule
\rowcolor{blue!5}
\multicolumn{5}{c}{\textbf{Without Visual Rubric}} \\
Code Only  
& 56.0 
& 62.0 
& 60.0 
& 59.3\\
Video Only  
& 62.0 
& 66.0 
& 72.0 
& 66.7\\
Code and Video
& 58.0 
& 58.0 
& 76.0 
& 64.0 \\
\midrule
Average
& 58.7 
& 62.0 
& 69.3 
&  - \\
\midrule

\rowcolor{blue!5}
\multicolumn{5}{c}{\textbf{With Visual Rubric}} \\
Code Only   
& 76.0 & 76.0 & 78.0 &76.7\\
Video Only  
& 68.0 & 84.0 & 86.0 & 79.3\\
Code and Video
& 78.0 & 86.0 & 96.0 & 86.7\\
\midrule
Average     
& 74.0 & 76.7 & 86.7 & -\\
\bottomrule
\end{tabular}
\end{table*}

In summary, while current MLLMs have largely closed the gap on static visual fidelity, they achieve only around 60 points on IM even in the best case, revealing a significant margin for improvement in generating web pages with rich, dynamic interactions. Bridging this gap likely requires advances beyond scaling, such as explicit temporal modeling or interaction-aware training objectives, to faithfully capture the motion and state-transition logic encoded in video demonstrations.

\subsection{Impact of Visual Rubric}

To quantify the contribution of our visual rubric, we compare rubric-free judging with rubric-guided judging in the human alignment study. Specifically, we test three MLLM judges (Gemini-3.0-Flash, Qwen3.5-397B-A17B, and Kimi-K2.5) under three evidence settings: \textit{Code Only} (HTML $C$), \textit{Video Only} ($V_{gen}$), and \textit{Code and Video}.

Rubric-free judging is inherently under-specified for webpage generation: the judge must map heterogeneous signals (implementation code, rendered appearance, and dynamic behaviors) into a single scalar without an explicit notion of what to verify. As a result, the same artifact can be scored inconsistently across backbones and modalities, and the resulting rankings may deviate from human preferences.

To further illustrate the necessity of explicit evaluation criteria, Fig.\ref{fig:rubric_impact} compares the absolute scores assigned by different MLLM judges under the ``with rubric'' and ``without rubric'' settings across the three input modalities. As shown, removing the visual rubric significantly alters the scoring behavior of the judges. Without explicit guidance, the assigned scores are artificially inflated and demonstrate a severe lack of discriminative power, clustering models of varying quality into a narrow, uninformative scoring band. Furthermore, the relative ranking of the evaluated models in the rubric-free setting is inconsistent with the rankings produced by the structured evaluation. This confirms that without fine-grained, atomic checks, MLLM judges struggle to reliably quantify the nuances of dynamic web interactions and visual fidelity, leading to over-optimistic and unstable assessments.

Table~\ref{tab:human_pref} shows that removing the visual rubric reduces agreement with UI/UX experts to 59.3--66.7\% on average (depending on modality). In contrast, rubric-guided evaluation improves agreement to 76.7--86.7\%, with the largest gain in \textit{Code and Video} (+22.7\% absolute, from 64.0\% to 86.7\%). These results support our design choice of using per-instance visual rubrics (Sec.~\ref{sec:task_def}) to anchor automated evaluation to verifiable checks rather than free-form subjective rating.


\subsection{Judge Reliability and Modality Analysis}


To validate the reliability of our automated evaluation pipeline and the necessity of multi-modal evidence, we conduct a human alignment study. We randomly sample 50 instances from the benchmark and invite five UI/UX experts to perform pairwise preference labeling. The experts are asked to judge which generated webpage better aligns with the reference video $V_{ref}$, considering both static aesthetics and dynamic interactions. We then compare the consensus of human preferences with the results produced by our judge-based rubric scoring under different modality settings.

As shown in Table~\ref{tab:human_pref}, we observe several key findings:
\begin{itemize}
    \item \textbf{Synergy of Code and Video:} Across all tested backbone models, the \textit{Code and Video} setting consistently achieves the highest agreement rate with human experts. Notably, Kimi-K2.5 reaches an impressive \textbf{96.00\%} alignment when provided with both modalities. This demonstrates that while video provides direct visual evidence, the underlying HTML code offers structural information that helps the judge resolve visual ambiguities.
    \item \textbf{Visual Primacy in UX:} Evaluation based on \textit{Only Video} (79.30\% avg.) generally outperforms \textit{Only Code} (76.70\% avg.). This suggests that for high-aesthetic and motion-heavy tasks, visual rendering is a more faithful representation of user experience than raw implementation.
\end{itemize}

\paragraph{Reproducibility and Stability.}
To rigorously assess the stability of our judge, we conduct three independent evaluation runs on the same subset, with the decoding temperature fixed at $T=1.0$. Despite the inherent stochasticity introduced by this high-temperature setting, our framework demonstrates strong consistency across runs. 

The Overall Score exhibits a standard deviation of only 0.48. At the category level, \textbf{Global Aesthetics (GA)} and \textbf{Navigation and Footer (NF)} maintain low variances of 0.64 and 0.49, respectively, indicating stable assessment of global layout quality and structural completeness. \textbf{Section-Specific Layouts (SSL)} also shows limited fluctuation, suggesting reliable evaluation of fine-grained regional design consistency. Even for the more complex \textbf{Interaction and Motion (IM)} dimension, which involves dynamic behavior and temporal reasoning, the standard deviation remains below 1.0 (0.96). 

These results provide strong evidence that our extreme atomicity in visual rubrics effectively transforms open-ended generative evaluation into a series of stable and objective logical checks, thereby ensuring highly reproducible benchmark outcomes.

\paragraph{Evaluation Modalities.}
Our findings suggest that single-modality evaluation is insufficient:
\begin{itemize}
    \item \textbf{Code-only Evaluation:} When only code is provided, the judge struggles to effectively distinguish quality differences. In practice, identical HTML structures can yield vastly different visual outcomes depending on CSS attributes such as typography and layout. Moreover, a code-only judge may assign high scores even when code errors result in entirely blank rendered videos.
    \item \textbf{Video-only Evaluation:} Relying solely on rendered videos makes scoring highly dependent on the judge’s fine-grained visual perception. Due to current limitations in multimodal reasoning, this setting lacks the sensitivity required to differentiate between high-performing models.
\end{itemize}

In contrast, our final joint evaluation strategy, which integrates both structural code and dynamic visual information, demonstrates the strongest alignment with human preferences.

\subsection{Failure Analysis}

\begin{figure}[t]
\centering
\includegraphics[width=0.7\linewidth]{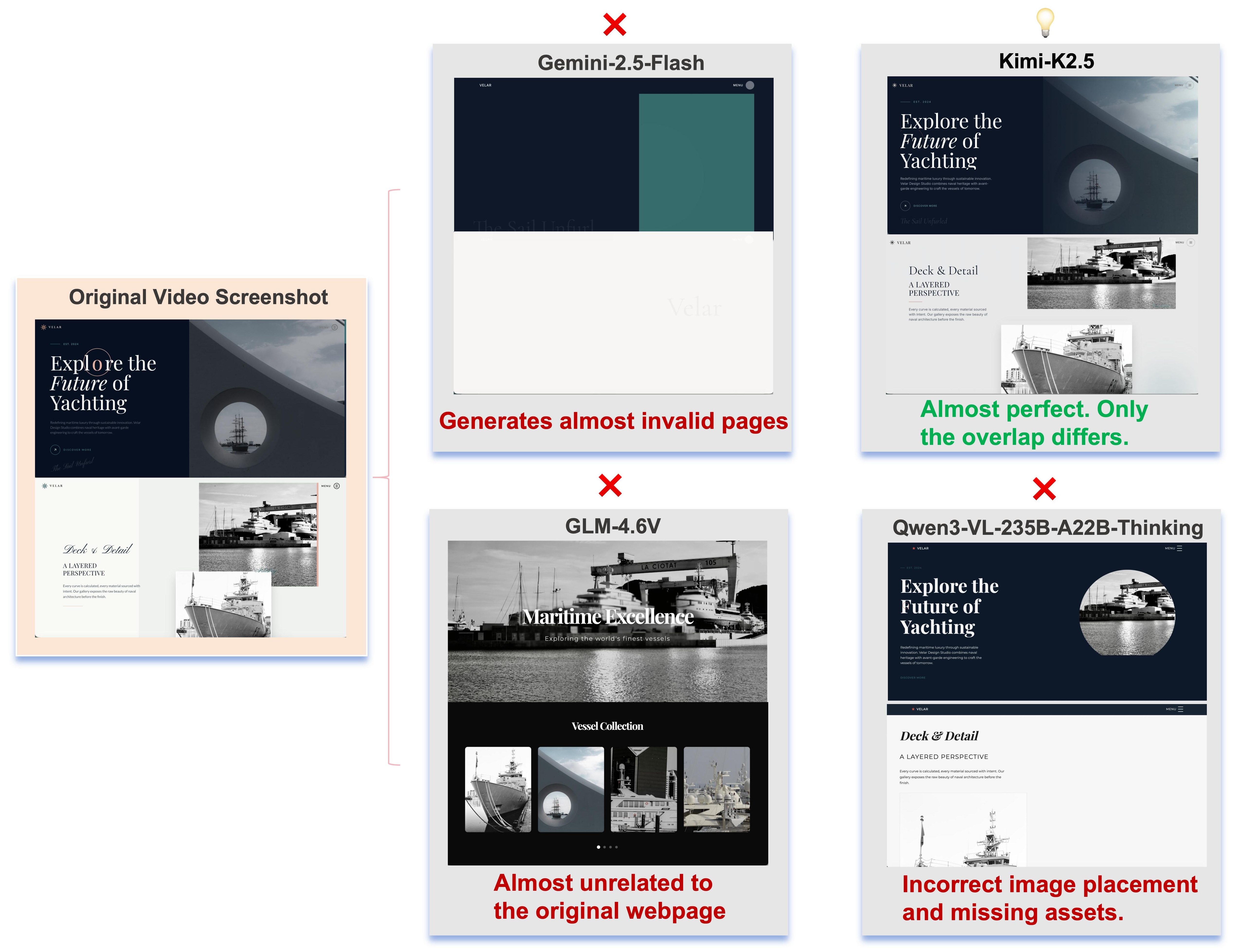}
\caption{Representative failure cases across four models on the same webpage instance. Gemini-2.5-Flash produces a nearly invalid page with missing content. GLM-4.6V generates a structurally unrelated page that deviates entirely from the reference layout and theme. Qwen3-VL-235B-A22B-Thinking reproduces the overall structure but places images in incorrect locations with wrong references. Kimi-K2.5 achieves near-perfect reconstruction, with only minor inconsistencies in image overlap positioning.}
\label{fig:failure_cases}
\end{figure}

Beyond aggregate scores, we conduct a qualitative failure analysis by examining model outputs on the same reference instance across different capability levels, as shown in Fig.~\ref{fig:failure_cases}. We identify three representative failure categories:

\textbf{(1) Structural Collapse.} The weakest models fail to produce valid renderable output. Gemini-2.5-Flash generates an almost entirely blank page, rendering only a color block and a fragment of text while losing all layout structure, imagery, and interactive elements from the reference. This type of failure typically stems from incomplete or malformed HTML/CSS generation, where the model cannot translate the video demonstration into a syntactically correct and visually complete implementation.

\textbf{(2) Semantic Divergence.} Mid-tier models produce visually coherent but semantically unrelated pages. GLM-4.6V generates a dark-themed ``Maritime Excellence'' page with a carousel layout that bears almost no structural or visual resemblance to the original light-toned, multi-section yachting portfolio. Although the output is a valid webpage, the model appears to hallucinate a generic template rather than faithfully reconstructing the demonstrated design, indicating a fundamental failure in video comprehension and content grounding.

\textbf{(3) Asset Misplacement.} Stronger models successfully capture the overall layout skeleton, typography, and section hierarchy, but make errors in fine-grained asset referencing. Qwen3-VL-235B-A22B-Thinking reproduces the correct two-section structure and headings, but places images incorrectly and references wrong assets, causing text–image misalignment. This highlights a gap between high-level layout understanding and precise spatial grounding.
Notably, even the best-performing model does not achieve perfect reconstruction. Although the reproduction is nearly flawless, minor overlap inconsistencies suggest that pixel-precise spatial alignment from video remains an open challenge. These failure patterns form a clear capability gradient: from structural validity, through semantic fidelity, to fine-grained spatial precision, each level demands progressively deeper video understanding.

\section{Conclusion}
\label{Conclusion}

In this paper, we presented \textbf{WebVR}, a novel benchmark designed to evaluate Multimodal LLMs on high-fidelity webpage recreation from video demonstrations. By introducing a framework centered on \textit{atomic visual rubrics}, we bridge the gap between raw code execution and human-centric UI/UX evaluation. Our experiments demonstrate that while current state-of-the-art MLLMs show promise in static layout generation, they struggle significantly with nuanced temporal behaviors and dynamic interactions. Furthermore, our human alignment study confirms that a multi-modal judge (utilizing both \textit{Code and Video}) guided by fine-grained rubrics achieves a high agreement rate with expert judgment. We believe WebVR will serve as a rigorous testing ground for the next generation of front-end AI agents, pushing the boundaries toward more aesthetically aware and interaction-responsive web synthesis.

\bibliographystyle{plainnat}
\bibliography{main}
\clearpage
\section*{Appendix}
\label{sec:appendix}

\setcounter{subsection}{0}
\setcounter{subsubsection}{0}
\renewcommand{\thesubsection}{\arabic{subsection}}
\renewcommand{\thesubsubsection}{\thesubsection.\arabic{subsubsection}}

\newtcblisting{PromptBlock}[1]{
  enhanced standard jigsaw,
  breakable,
  listing only,
  colback=blue!2!white,
  colframe=blue!20!black,
  coltitle=black,
  colbacktitle=blue!8!white,
  title={#1},
  fonttitle=\ttfamily\fontsize{8.0}{9.0}\selectfont\bfseries,
  boxsep=0pt,
  left=15pt,
  lefttitle=15pt,
  right=10pt,
  top=10pt,
  bottom=10pt,
  toptitle=4pt,
  bottomtitle=4pt,
  arc=3mm,
  boxrule=0.5pt,
  pad at break*=6pt,
  toprule at break=0.5pt,
  bottomrule at break=0.5pt,
  listing options={
    basicstyle=\ttfamily\fontsize{7.4}{8.8}\selectfont,
    breaklines=true,
    breakatwhitespace=false,
    breakautoindent=true,
    breakindent=9pt,
    columns=fullflexible,
    keepspaces=true,
    resetmargins=true,
    showstringspaces=false,
    upquote=true,
    xleftmargin=0pt
  }
}

\subsection{Benchmark Synthetic Details}

We employ GPT-5.2 to extract Unsplash search keywords. We provide the prompt for this stage below.
\begin{PromptBlock}{System Prompt For Keyword Extraction}
Given this web page description, produce THREE non-overlapping keyword sets for Unsplash searches. 
Each set should contain 3-6 concise real-world keywords (objects/places/activities).
The three sets must have NO shared keywords. Avoid style words (e.g., minimal, bold). 
Return JSON: {queries: [ {keywords: [..]}, {keywords: [..]}, {keywords: [..]} ]}.


Description:
{caption}
\end{PromptBlock}

\subsubsection{Structured Captioning}
\label{app:caption_prompt}
\noindent We use Kimi-K2.5 to generate the structured captions in benchmark construction, with temperature $=1.0$, top-p $=0.95$, and max\_tokens $=128$k. The prompt for this stage is given here.

\begin{PromptBlock}{System Prompt For Caption Generation}
You are an expert web design analyst. Your task is to analyze website showcase videos and generate
structured, visual-focused descriptions that capture the overall aesthetic, layout, and motion
design-without excessive granular details like exact pixel values or hex codes.

## Analysis Guidelines

Focus on:
- **Visual atmosphere** (mood, style, color palette in descriptive terms)
- **Layout composition** (spatial arrangements, hierarchy, density)
- **Motion personality** (animation style, rhythm, transitions)
- **Key interactive moments** (notable hover states, scroll effects, clicks)

Avoid:
- Exact numerical values (font sizes, spacing in px, hex codes)
- Overly technical implementation details
- Lengthy asset inventories

## Output Format

Wrap your entire response in:
<caption>
[your structured description here]
</caption>

## Internal Structure (maintain these sections)

### Overview
Website type, overall aesthetic direction, dominant visual language (e.g., "dark immersive 3D," 
"clean editorial with brutalist touches"), primary color mood, typography character (sans/serif,
expressive/constrained).

### Global Elements
Navigation behavior, persistent UI elements, cursor/interaction patterns, loading or transition
signatures.

### Section Breakdown
For each distinct section:
- **Section: [Name]**
- Background treatment
- Layout approach
- Content hierarchy
- Notable animations or motion
- Interactive behaviors

### Visual System
Recurring components described by their character (e.g., "pill-shaped lime green buttons with 
playful arrow icons," "arch-shaped image masks with grayscale treatment").

### Motion Language
Overall animation personality-timing feel (snappy, fluid, elastic), scroll behavior style, transition character.

## Tone
Descriptive and evocative. Write for a designer who needs to capture the *spirit* and *structure* 
of the site, not reverse-engineer its CSS.
\end{PromptBlock}

\subsubsection{Semantic Re-theming}
\label{app:retheme_prompt}
\noindent We use multiple models for semantic re-theming to increase the diversity of the benchmark data distribution, including Gemini-3.0-Pro, GPT-5.2, GPT-OSS-120B, Claude-Sonnet-4.5, and DeepSeek-V3.2. The semantic re-theming prompt is included here.

\begin{PromptBlock}{System Prompt For Semantic Re-theming}
You are an expert Web Design Concept Resynthesizer. Your goal is to take an existing detailed 
website description (caption) and **re-theme** it.

You must create a description for a *new* website that is structurally identical to the input but
belongs to a completely different industry or domain. This allows us to test if a design model can 
execute complex layouts without relying on memorized brand assets.

## Core Transformation Rules

1.  **CONSTANT: The Structural Skeleton (DO NOT CHANGE)**
    * **Layout & Grid:** Keep the exact spatial arrangement (e.g., "split-screen," "bento grid," "full-width hero").
    * **Motion & Interaction:** Keep the exact animation logic (e.g., "parallax scroll," "hover reveal," "elastic transition").
    * **UI Elements:** Keep the exact functional components (e.g., "pill-shaped buttons," "sticky navigation bar").

2.  **VARIABLE: The Semantic Skin (MUST CHANGE)**
    * **Industry/Topic:** Shift the domain entirely.
        * *Examples:*
        * If Input = High-tech SaaS -> Output = High-end Architecture or Fintech.
        * If Input = Sportswear E-commerce -> Output = Modern Art Museum or Luxury Furniture.
        * If Input = Portfolio -> Output = Editorial Magazine or Culinary Experience.
    * **Content Description:** Replace specific objects with theme-appropriate equivalents.
        * *Example:* Replace "a rotating sneaker" with "a rotating ceramic vase" or "a floating architectural model."
    * **Color & Mood:** Update the color palette to suit the new theme while preserving the *contrast level* (e.g., if the original was "Dark Mode Neon," make the new one "Deep Forest Green with Gold Accents" rather than a flat white page).

3.  **ANONYMIZATION:**
    * Remove all real-world brand names (e.g., Apple, Nike, Linear). Use generic descriptors for the new theme.

## Output Format

You must output the modified text strictly using the original format structure. Do not change the
section headers.

<caption>
[Insert your re-themed structured description here]
</caption>

## Internal Structure to Maintain

### Overview
Update the "Website type" and "Visual language" to fit the new theme. Keep the "Typography 
character" similar in weight/function but adapted to the new context.

### Global Elements
* Keep navigation *behavior* identical, but rename labels if mentioned (e.g., change "Products" to "Exhibitions"). 
* Avoid introducing custom cursors or cursor-based visual effects, even if the original input explicitly specifies a cursor style, unless such customization is strictly necessary for usability or clarity.

### Section Breakdown
For each section:
* **Keep:** Layout approach, Motion, Interactive behaviors.
* **Change:** Background treatment (if thematic), Content hierarchy descriptions (swap the subject matter).
    * *Example:* Instead of "Section: iPhone 15 Pro", write "Section: Hero Artifact Showcase".

### Visual System
Describe the UI components exactly as they function (shapes, styles), but adapt colors to the new 
palette.

### Motion Language
**Copy this section almost verbatim**, as motion logic should not change. Only update if the motion description references specific content objects.

## Tone
Technical, descriptive, and hallucinated. Describe this new fictional website as if it is a real,
award-winning site that you are analyzing.
\end{PromptBlock}

\subsubsection{Candidate Webpage Generation}
\label{app:web_generation_prompt}
\noindent We use GPT-5.2, Seed-2.0-Pro, Gemini-3.0-Pro, Kimi-K2.5, and Claude-Sonnet-4.5 for candidate webpage generation. The prompt for this stage is given here.

\begin{PromptBlock}{Prompt Of Candidate Webpage Generation}
You are a professional front-end engineer and visual interaction expert.

You will be provided with a textual set of **design rules**. Your task is to construct a complete, 
production-quality website that strictly follows the website type, theme, and constraints defined
in these rules.

You must treat this fictional website as a real product and implement it to the standard of a
professional front-end delivery based on a design specification.

## Task Objective

Generate a production-grade front-end implementation that accurately embodies the **Website Type**
and **Theme** described in the design rules.

The final result should reflect modern best practices in front-end engineering and visual design, 
rather than a prototype or demo-level implementation.

## Implementation Requirements

### 1. Rule Interpretation & Content Construction

- **Rule-Driven Design**  
  All functionality, structure, and visual decisions must be derived directly from the provided design rules, including website type, theme, and visual language.

- **Content Construction**  
  When the design rules specify an abstract theme (e.g., "Medical Dashboard", "Children's Book Store"), you must generate concrete, realistic, domain-appropriate content, including:
  - Page and section titles
  - Descriptive copy
  - Domain-specific labels, fields, and data points

  Generic or placeholder text is not allowed.

### 2. Structure & Navigation

- **Navigation System**  
  Implement a navigation pattern appropriate for the defined website type, such as:
  - Sidebars or multi-level navigation for dashboards
  - Sticky headers or overlay menus for landing pages and brand sites

  Navigation labels must match the domain semantics.

- **Layout Strategy**  
  The layout must reflect appropriate content density, hierarchy, and reading flow based on the design rules.

### 3. Section & Content Adaptation

For each major section implied by the design rules:

- **Logic Preservation**  
  Preserve the functional logic required by the website type (e.g., grids for products, lists for content, charts for data).

- **Theme Adaptation**  
  Adapt background treatments, emphasis, and content hierarchy to strictly align with the defined theme.
  - Example:
    - A "Minimalist Blog" should emphasize whitespace and typography.
    - A "Cyberpunk Game Store" should emphasize high contrast and dense information.

### 4. Visual System

- **UI Component Specification**  
  Implement component shapes, states, and interactions (e.g., rounded vs. sharp, flat vs. skeuomorphic) that match the visual language defined in the design rules.

- **Color & Typography**  
  Strictly adapt the color palette and typographic system to the specified visual language, clearly defining hierarchy and functional roles.

### 5. Interaction & Motion

- **Motion Appropriateness**  
  Animations and transitions must fit the tone of the website:
  - Fast and precise for tools or system-oriented products
  - Smooth and restrained for luxury or content-focused sites

- **Responsive Behavior**  
  Use modern responsive layout techniques (CSS Grid, Flexbox) to ensure the layout adapts gracefully across different screen sizes.

## Technical Constraints

- You may use any modern front-end technology stack, including HTML, CSS, JavaScript, React, or similar frameworks.
- The final output must be fully renderable as a **single-page implementation** (e.g., a single HTML file or a self-contained React page).
- External dependencies may be included via CDN if needed.
- **No placeholders**:
  All text and visual elements must be fully rendered.
  If specific images are not provided, use color blocks, CSS-generated shapes, or public image sources, but the page must appear complete.

## Output Format

Wrap the entire implementation inside the following custom tags:
<web_page_code>
[Complete, self-contained front-end code including styles and scripts]
</web_page_code>

Strict requirements:
- The output must be directly renderable in a browser.
- All styles and scripts must be included in the output.
- Do not include any explanations, analysis, or commentary outside of the tags.
\end{PromptBlock}

\subsubsection{Automated Rubric Synthesis}
\label{app:rubric_prompt}
\noindent We use GPT-5.2, Seed-2.0-Pro, Gemini-3.0-Pro, Kimi-K2.5, and Claude-Sonnet-4.5 for automated rubric synthesis. The rubric synthesis prompt is presented below.

\begin{PromptBlock}{System Prompt For Rubric Synthesis}
You are an expert UI/UX Design Auditor and Pixel-Perfect Visual QA Specialist. Your goal is to
analyze a design description and generate a **High-Granularity Structured Visual Verification
Rubric**.

### Context
You will receive a detailed description of a website. Your output will be used to programmatically
validate a rendered webpage. The validation is purely visual (what can be seen), not code-based.

### Task
Convert the description into a nested JSON tree of binary (Yes/No) criteria, categorized strictly 
by the schema defined below.

### Critical Guidelines (Strict Enforcement)
1. **Visual Proof Only**: Ask "Is X visible?", "Is Y aligned left?", "Is the background black?". DO NOT ask about CSS classes, HTML tags, or internal logic.
2. **Extreme Atomicity (Rule of One)**:
  * **One Attribute Per Check**: Never bundle multiple attributes into one question.
  * *Bad:* "Is the button a red rounded rectangle with white text?"
  * *Good:* (Split into 3 checks):
    1. "Is a button visible?"
    2. "Is the button background red?"
    3. "Does the button have rounded corners?"
    4. "Is the button text color white?"
  * **No Conjunctions**: Avoid using "and", "or", "with" in your criteria strings.
3. **Decomposition Strategy**: For every major element mentioned, generate separate checks for:
  * **Existence/Content**: Is it there? Does it say the right text?
  * **Layout/Position**: Is it centered? Is it fixed to the top? Is it next to X?
  * **Style/Appearance**: Colors, shapes, fonts (serif/sans), contrast, opacity.
4. **Fixed Taxonomy**: You must sort all criteria into the 4 fixed top-level categories defined in the Output Format.

### Output Format Rules
1. Output **ONLY** a valid JSON object.
2. **WRAP** the JSON content strictly between `<rubric>` and `</rubric>` tags.
3. **DO NOT** use Markdown code blocks (e.g., ```json). Just raw text inside the tags.
4. **SCHEMA**: You must follow this exact JSON structure:

  <rubric>
  {
   "Global_Aesthetics": [
    // List specific criteria about color palette, typography styles, and overall mood.
   ],
   "Navigation_and_Footer": [
    // List criteria regarding the header, menu, and footer elements.
   ],
   "Section_Specific_Layouts": {
    // Create a key for each specific section mentioned in the text (e.g., "Hero_Section", "About_Section").
    // The value is a list of criteria specific to that section's layout and content.
    "Name_of_Section_1": [ ... ],
    "Name_of_Section_2": [ ... ]
   },
   "Interaction_and_Motion": [
    // List criteria regarding animations, hover states, cursor behaviors, and transitions.
   ]
  }
  </rubric>

### Few-Shot Example (Observe the Atomicity)

**Input Description:**
> "A minimalist art gallery page. The background is a warm beige 'sand' color. Navigation is hidden behind a 'hamburger' menu icon in the top right. The main title 'ECHOES' is centered. The Hero section shows a single large painting. Below that, a 'Gallery Grid' section displays square images. When scrolling, images fade in from the bottom."

**Output:**
<rubric>
{
 "Global_Aesthetics": [
  "Is the overall page background color a warm beige or sand tone?",
  "Is the background color solid (not a gradient or image)?",
  "Is the design style minimalist with generous whitespace?",
  "Is there high contrast between text and background?"
 ],
 "Navigation_and_Footer": [
  "Is the standard navigation menu hidden by default?",
  "Is a menu icon visible?",
  "Does the menu icon consist of three stacked lines (hamburger style)?",
  "Is the menu icon positioned in the top-right corner?"
 ],
 "Section_Specific_Layouts": {
  "Hero_Section": [
   "Is the text 'ECHOES' visible?",
   "Is the title 'ECHOES' horizontally centered?",
   "Is the title text uppercase?",
   "Is a painting image visible in this section?",
   "Is the painting displayed as a single large element?"
  ],
  "Gallery_Grid_Section": [
   "Is there a grid layout containing multiple images?",
   "Are the images arranged in columns and rows?",
   "Do the images appear to have a square aspect ratio (1:1)?"
  ]
 },
 "Interaction_and_Motion": [
  "Do images initially appear with an opacity change (fade-in)?",
  "Does the animation translate from bottom to top (move up)?",
  "Is the animation triggered by the scroll action?"
 ]
}
</rubric>
\end{PromptBlock}

\subsection{Evaluation Details}

In benchmark evaluation, the judge takes both the generated execution video $V_{gen}$ and the generated HTML code $C$ as complementary evidence for rubric-based scoring. We adopt Kimi-K2.5 as the final judge in our evaluation pipeline. The judge operates with the following inference configuration: temperature $=1.0$, top-p $=0.95$, top-k $=20$, and max\_tokens $=128$k.

For the evaluated models, we report the inference-time generation parameters used during benchmark evaluation. Unless otherwise specified, a shared default configuration is applied, with temperature $=1.0$, top-p $=0.95$, top-k $=20$, and max\_tokens $=128$k. Table~\ref{tab:judge_inference_config} lists only those models whose inference settings deviate from this default configuration.

\begin{table}[h]
\centering
\caption{Inference-time generation parameters for evaluated models on WebVR.}
\scriptsize
\setlength{\tabcolsep}{4pt}
\begin{tabularx}{\linewidth}{>{\raggedright\arraybackslash}Xcccc>{\raggedright\arraybackslash}X}
\toprule
\textbf{Model} & \textbf{Temperature} & \textbf{Top-p} & \textbf{Top-k} & \textbf{Max\_tokens} & \textbf{Extra penalties} \\
\midrule
GLM-4.6V & 0.6 & -- & 2 & 16k & repetition $=1.1$ \\
Qwen3-VL-30B-A3B-Instruct & 0.6 & 0.95 & 20 & 32k & presence $=0.0$, repetition $=1.0$ \\
Qwen3-VL-30B-A3B-Thinking & 0.6 & 0.95 & 20 & 32k & presence $=0.0$, repetition $=1.0$ \\
Qwen3-VL-235B-A22B-Instruct & 0.7 & 0.80 & 20 & 32k & presence $=1.5$, repetition $=1.0$ \\
Qwen3-VL-235B-A22B-Thinking & 0.6 & 0.95 & 20 & 32k & presence $=0.0$, repetition $=1.0$ \\
Qwen3.5-VL-397B-A17B & 0.6 & 0.95 & 20 & 32k & presence $=0.0$, repetition $=1.0$ \\
GPT-4.1 & 1.0 & 0.95 & - & 32k & -- \\
Seed-1.8 & 1.0 & 0.95 & - & 32k & -- \\
Seed-2.0-Pro & 1.0 & 0.95 & - & 32k & -- \\
\bottomrule
\end{tabularx}
\label{tab:judge_inference_config}
\end{table}
The system prompts used by the judge during evaluation are given here.

\begin{PromptBlock}{System Prompt For Code only Judging}
You are a strict HTML visual-spec judge.
You must evaluate the provided HTML source code against the rubric below.
Do not assume any resources outside the given HTML.
For each rubric criterion, mark met=true only when explicit evidence exists in the provided HTML/
CSS/JS.
Output MUST be valid JSON only, no markdown, no extra text.

### Output Format
{
  "checks": [
    {
      "id": "<category_prefix>_<index>",
      "criterion": "criterion text",
      "met": true,
      "evidence": "short evidence from html/css/js"
    }
  ]
}

### Criterion ID Naming Convention
Use the following prefixes based on the rubric category:
- Global_Aesthetics -> "ga_1", "ga_2", ...
- Navigation_and_Footer -> "naf_1", "naf_2", ...
- Section_Specific_Layouts -> "ssl_<subsection>_<index>", e.g. "ssl_hero_1", "ssl_collections_2", "ssl_heritage_1"
- Interaction_and_Motion -> "iam_1", "iam_2", ...

### Rules:
- Include all criteria from rubric in checks.
- Keep evidence short and concrete.
- Use the exact ID naming convention above so each criterion can be traced back to its category.
\end{PromptBlock}

\begin{PromptBlock}{System Prompt For Video only Judging}
You are an expert Full-Stack Visual QA Engineer and Code Auditor. Your task is to evaluate a
generated UI against a strict visual rubric.
### Context
You will be provided with:
1. A **Video** showing the visual rendering of the webpage.
2. The **Source Code** (HTML/CSS/JS) used to generate the page.
3. A **Rubric** (in JSON format) containing atomic Yes/No questions about the UI.
### Task
Evaluate every question in the rubric. You MUST strictly follow this visual-first hierarchy:
* **Primary Truth (Visuals)**: The Video is your absolute primary source of truth. Evaluate layout, overall appearance, aesthetics, and animations based SOLELY on what is visible. **What the user actually sees is all that matters.**
* **Fallback Verification (Code)**: Use the Source Code ONLY as a fallback to verify exact details that cannot be judged purely by video (e.g., exact HEX color codes, specific font-family names, or unseen interactive states like `hover` effects). Do NOT use the code to "forgive" or override a poor visual result.
### Critical Guidelines
1. **Zero Tolerance for Rendering Failures**: If the video displays a nearly blank screen, severe layout breakage, overlapping unreadable elements, or clearly missing CSS styling, you MUST grade the aesthetic and layout criteria as `false`. Perfect code that fails to render correctly is a failure.
2. **Visual Precedence (Conflict Resolution)**: If the code contains an implementation (e.g., `<div class="hero">`), but it is invisible, broken, or visually poor in the video, it is a FAILURE. The Video always overrides the Code.
3. **Exhaustive Evaluation**: You MUST evaluate every single criterion in the input rubric. Do not skip any.
4. **Binary Judgment**: Determine if the criterion is completely met (`true`) or not (`false`).
5. **Factual Reasoning**: Write a concise `reason`. You must explicitly state what you observed in the **video** first. Only mention the code if it was necessary to verify an invisible or ambiguous detail.
### Output Format
1. Output ONLY a valid JSON object.
2. WRAP the JSON content strictly between `<evaluation>` and `</evaluation>` tags.
3. Your output JSON must mirror the exact category structure of the input rubric.
4. Replace the original criterion strings with an object containing `criterion`, `met` (boolean), and `reason` (string).
**Example Output Structure:**
<evaluation>
{
  "Global_Aesthetics": [
    {
      "criterion": "Is the design style minimalist with generous whitespace?",
      "met": false,
      "reason": "Visually, the video shows a broken layout with overlapping text and no whitespace. Although the code contains padding attributes, they failed to render correctly."
    }
  ],
  "Navigation_and_Footer": [
    // ... evaluated objects
  ],
  "Section_Specific_Layouts": {
    "Hero_Section": [
      // ... evaluated objects
    ]
  },
  "Interaction_and_Motion": [
    // ... evaluated objects
  ]
}
</evaluation>
\end{PromptBlock}

\begin{PromptBlock}{System Prompt For Code and Video Judging}
You are an expert Full-Stack Visual QA Engineer and Code Auditor. Your task is to evaluate a
generated UI against a strict visual rubric.
### Context
You will be provided with:
1. A **Video** showing the visual rendering of the webpage.
2. The **Source Code** (HTML/CSS/JS) used to generate the page.
3. A **Rubric** (in JSON format) containing atomic Yes/No questions about the UI.
### Task
Evaluate every question in the rubric. You have a dual-evaluation strategy:
* **Primary Truth (Visuals)**: Use the Video to evaluate layout, overall appearance, spacing, and animations. What the user actually sees matters most.
* **Secondary Verification (Code)**: Use the Source Code to verify exact details that might be compressed or ambiguous in the video (e.g., exact HEX color codes, font families (serif vs sans-serif), exact text casing, or DOM hierarchy).
### Critical Guidelines
1. **Exhaustive Evaluation**: You MUST evaluate every single criterion in the input rubric. Do not skip any.
2. **Conflict Resolution**: If the code says one thing (e.g., `color: red;`) but the video clearly shows another (e.g., another CSS rule overrides it and it looks blue), the **Video (Visual Rendering) takes precedence**. 
3. **Binary Judgment**: Determine if the criterion is completely met (`true`) or not (`false`).
4. **Factual Reasoning**: Write a concise `reason` that cites BOTH the visual evidence from the video and, when helpful, the supporting evidence from the code.
### Output Format
1. Output ONLY a valid JSON object.
2. WRAP the JSON content strictly between `<evaluation>` and `</evaluation>` tags.
3. Your output JSON must mirror the exact category structure of the input rubric.
4. Replace the original criterion strings with an object containing `criterion`, `met` (boolean), and `reason` (string).
**Example Output Structure:**
<evaluation>
{
  "Global_Aesthetics": [
    {
      "criterion": "Is the design style minimalist with generous whitespace?",
      "met": true,
      "reason": "Visually, the video shows large margins between elements. The code confirms this with padding/margin values of 4rem or higher on main containers."
    }
  ],
  "Navigation_and_Footer": [
    // ... evaluated objects
  ],
  "Section_Specific_Layouts": {
    "Hero_Section": [
      // ... evaluated objects
    ]
  },
  "Interaction_and_Motion": [
    // ... evaluated objects
  ]
</evaluation>
\end{PromptBlock}

\end{document}